\documentclass{article}
\usepackage[round]{natbib}
\usepackage[preprint]{neurips_2024}
\usepackage[utf8]{inputenc} 
\usepackage[T1]{fontenc}    
\usepackage{hyperref}       
\usepackage{url}            
\usepackage{booktabs}       
\usepackage{amsfonts}       
\usepackage{nicefrac}       
\usepackage{microtype}      
\usepackage{lipsum}
\usepackage{fancyhdr}       
\usepackage{graphicx}       
\graphicspath{{media/}}     
\usepackage{amsmath}
\usepackage{algorithm}
\usepackage{algpseudocode}
\usepackage{graphicx}
\usepackage{subcaption} 
\usepackage[table]{xcolor}
\usepackage{multirow}
\usepackage[most]{tcolorbox} 

\title{MOTIF: Modular Thinking via Reinforcement Fine-tuning in LLMs
}

\author{
  Purbesh Mitra \\
  University of Maryland \\
  \texttt{pmitra@umd.edu} \\
   \And
  Sennur Ulukus \\
  University of Maryland \\
  \texttt{ulukus@umd.edu} \\
}

\begin{document}
\maketitle

\begin{abstract}
Recent advancements in the reasoning capabilities of large language models (LLMs) show that employing group relative policy optimization (GRPO) algorithm for reinforcement learning (RL) training allows the models to use more thinking/reasoning tokens for generating better responses. However, LLMs can generate only a finite amount of tokens while maintaining attention to the previously generated tokens. This limit, also known as the context size of an LLM, is a bottleneck in LLM reasoning with arbitrarily large number of tokens. To think beyond the limit of context size, an LLM must employ a modular thinking strategy to reason over multiple rounds. In this work, we propose \textbf{MOTIF: Modular Thinking via Reinforcement Finetuning} -- an RL training method for generating thinking tokens in multiple rounds, effectively allowing the model to think with additional context size. We trained the open-source model Qwen2.5-3B-Instruct on GSM8K dataset via parameter efficient fine-tuning and tested its accuracy on MATH500 and AIME2024 benchmarks. Our experiments show 3.8\% and 3.3\% improvements over vanilla GRPO based training in the respective benchmarks. Furthermore, this improvement was achieved with only 15\% of samples, thus demonstrating sample efficiency of MOTIF. Our code and models are available at \url{https://github.com/purbeshmitra/MOTIF} and \url{https://huggingface.co/purbeshmitra/MOTIF}, respectively.
\end{abstract}


\section{Introduction}\label{sec: intoduction}
In recent studies, large language models (LLMs) have been shown to demonstrate reasoning capabilities by training with reinforcement learning with verifiable rewards (RLVR). With the publication of Deepseek-R1~\citep{guo2025deepseek} and other subsequent studies~\citep{chen2025towards}, a relation between the number of LLM inference tokens and the reasoning accuracy has been established. With more number of tokens used in thinking, the LLMs are able to generate their long intermediate chain-of-thoughts~\citep{wei2022chain} automatically, thereby, resulting in more accurate answers. This new paradigm of LLM training has been referred to as test time compute or inference time compute~\citep{snell2024scaling}. This new research opened up a plethora of new avenues in the research community to discover simple, scalable algorithms that incentivize LLM reasoning by longer responses during inference. Even with non-reasoning LLMs, there is a direct relation between more tokens and better output, as shown in the works, like mixture-of-agents~\citep{wang2024mixture, li2024smoa}, where the responses from different LLMs were appended to get better response. Works by \cite{li2025rethinking, mitra2024distributed} showed that using a single LLM with high temperature creates diverse answers to put in a long context for better collaborative inference. A similar idea was employed by \cite{muennighoff2025s1}, which shows how test time compute can be scaled very simply by appending the reasoning tokens generated parallelly by the LLM with words, like ``Wait'' to improve the LLM performance.

However, all these methods are constrained by the context size of the LLM. Since an LLM is only able to calculate attention scores over only a finite number of past tokens, the reasoning capabilities are strictly bottlenecked by the context size. Furthermore, it has been observed that the quality of attention and information propagation over generated tokens declines as the number of tokens increases~\citep{liu2023lost}. To overcome this bottleneck, proprietary reasoning LLMs, like OpenAI-o1~\citep{jaech2024openai}, and Google-Gemini~\citep{team2023gemini}, provide very large (at least 128k tokens) context size. Indeed, calculating attention efficiently over a large number of tokens is an active area of research. \cite{munkhdalai2024leave} has shown that attention can be increased over one million tokens. \cite{ye2025infinite} proposes an attention allocation mechanism that facilitates information retrieval accurately over arbitrarily many tokens. Very recently, some works~\citep{yan2025inftythink,tian2025think, radha2024iteration, goldie2025synthetic, ning2025not} have proposed a multi-round thinking architecture with LLMs for long context reasoning. \cite{yan2025inftythink} has proposed an architecture that requires distilled data set for supervised fine-tuning to enable long-context reasoning. \cite{goldie2025synthetic} has proposed a stepwise reward method that uses process filtered data and outcome filtered data for assigning rewards in each round. These methods contrast with the simple rule based function proposed in Deepseek-R1~\citep{guo2025deepseek}. \cite{ning2025not} has proposed a dual LLM system for multi-turn reasoning that incorporates short and long chain of thoughts with multi-turn reinforcement learning. To the best of our knowledge, there has not been any work in the literature that addresses RLVR in multi-round LLM reasoning with purely outcome based reward function.

\begin{figure}
    \centering
    \includegraphics[width=\linewidth]{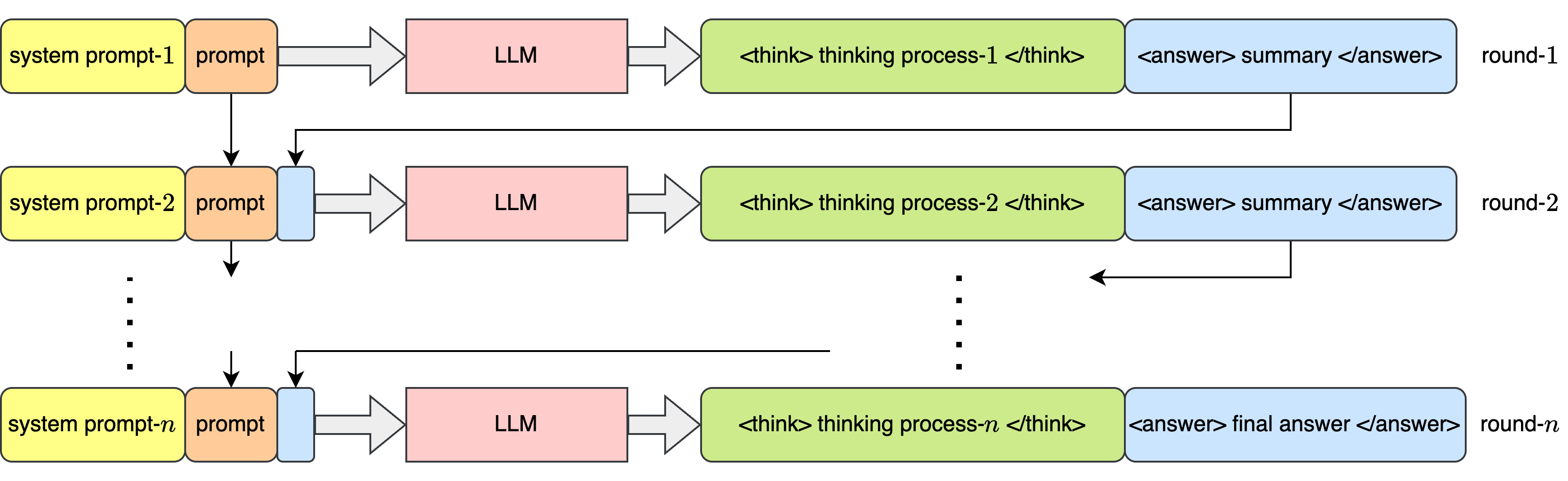}
    \caption{Multi-round inference for long context reasoning.}
    \label{fig: multiround}
\end{figure}

To that regard, we propose \textbf{MOTIF}: \textbf{\underline{Mo}}dular \textbf{\underline{T}}hinking via Re\textbf{\underline{i}}nforcement \textbf{\underline{F}}ine-tuning -- a reinforcement learning formulation to train multi-round LLM inference for long context reasoning. The architecture we consider, as shown in Fig.~\ref{fig: multiround}, is similar to the INFTYTHINK architecture proposed by \cite{yan2025inftythink}. A prompt is passed through an LLM in multiple rounds. In each round, the LLM generates, thinking tokens and answer tokens. The difference from vanilla reasoning LLM here is that the LLM does not try to generate a final answer in the answer tokens in the first pass. Rather, the system prompt of the LLM instructs it to do only partial progress on solving the question in the prompt. The LLM then generates a summary of its thinking that includes any plan or partial progress it has made. From second round onwards, the answer from the last round is fed along with the prompt for further thinking and answering. In the final round, the LLM is asked to generate a final answer. In this work, we formulate a simple rule based reward for training, such a system with reinforcement learning. Our key idea is to let the LLM generate some responses in the first round. Now for each of the responses, we generate multiple trajectories for the next few rounds and get the final answer from the LLM. Then, we estimate a probability of getting the correct answer from these trajectories and get the future accuracy reward from comparing with the correct solution. This future accuracy, reward, along with the format reward of the first round is used to train the LLM with group relative policy optimization (GRPO)~\citep{shao2024deepseekmath} algorithm. We train the open-source models Qwen2.5-3B-Instruct on questions from GSM8K dataset~\citep{cobbe2021training} via low rank adaptation (LoRA)~\cite{hu2022lora} -- a parameter efficient fine-tuning method and test the pass@1 accuracy on MATH500~\citep{hendrycks2021measuring, lightman2023let} and AIME24~\citep{li2024numinamath} benchmarks. Our results indicate strong performance boost in both the benchmarks. In MATH500 benchmark, MOTIF trained model accuracy is 48.6\%, compared to 44.8\% accuracy of vanilla GRPO trained model. In AIME2024 benchmark, MOTIF trained model scored 6.67\% accuracy compared to 3.33\% accuracy of vanilla GRPO trained model. Additionally, these performance improvements of MOTIF were achieved by training the model with only 15\% of samples used in the GRPO training. This indicates that MOTIF is more sample efficient than GRPO.

Our contributions are summarized follows:
\begin{itemize}
    \item We introduce MOTIF: Modular Thinking via Reinforcement Fine-tuning -- a method of training LLMs via reinforcement learning for long context reasoning tasks in a multi-round inference architecture.
    \item MOTIF utilizes an outcome based reward function for its GRPO pipeline, thus eliminating any need for process supervision in the intermediate reasoning steps.
    \item MOTIF shows significant improvement in pass@1 accuracy over vanilla GRPO training. This has been validated by testing MOTIF on MATH500 and AIME24 benchmarks, which shows 3.8\% and 3.3\% improvement, respectively.
    \item The performance improvement by MOTIF was achieved by training it with only 15\% samples of vanilla GRPO based training, thus demonstrating its sample efficiency.
\end{itemize}

\section{Background and Related Works}\label{sec: background}
The idea of solving a reasoning based task, iteratively over multiple rounds, was first proposed by \cite{radha2024iteration}. Their idea of \emph{iteration of thought} tries to simulate an inner reasoning dialogue of LLM, which adapts dynamically based on the evolving context of the prompt query and the response of the LLM. This reasoning dialogue is generated in multiple inference rounds by two different agents--an inner dialogue agent for generating prompts and an LLM agent for detailed response. This method shows performance gain compared to regular Chain of Thought~\citep{wei2022chain} or Tree of Thought~\citep{yao2023tree} methods.

In the post-DeepSeek era, \cite{yan2025inftythink} introduced INFTYTHINK, an architecture for iterative reasoning, thereby, extending the context size arbitrarily. INFTYTHINK transforms reasoning into a multi-round inference process with intermediate summarization; similar to our inference model, as shown in Fig.~\ref{fig: multiround}. This approach of interleaving reasoning segments with progress summaries allows for theoretically unbounded reasoning depth while maintaining bounded computational costs. The authors also present a method for reconstructing existing long-context reasoning datasets (demonstrated with OpenR1-Math) into this iterative
format for supervised fine-tuning. A similar architecture was proposed by \cite{tian2025think}, where the LLM is prompted in multiple rounds to answer a question and in each round, it is asked to re-answer based on the answer of the previous round. This approach does not require any training and shows improvement over single-round reasoning.

Training an LLM over multi-round inference was first shown by \cite{goldie2025synthetic}. They introduced Step-Wise Reinforcement Learning (SWIRL), a methodology for improving multi-step reasoning and tool use of LLMs. This approach involves first generating synthetic multi-step trajectories where an LLM can use tools like search engines or calculators, and then learning from this data using a step-wise RL with a generative reward model that evaluates each action in context. Experiments show that SWIRL significantly outperforms baseline approaches on various multi-step tool use, question answering, and mathematical reasoning tasks, and importantly, exhibits generalization across different out of distribution tasks and datasets.

The work by \cite{ning2025not} proposes a synergizing-oriented multi-turn reinforcement learning framework designed to make LLM reasoning more efficient. This process involves two specialized models — a long-thought LLM for complex steps and a short-thought LLM for simpler ones — that collaborate through a multi-turn conversation to solve problems. These models are trained using an asynchronous, alternating optimization strategy, where one model's policy is updated while the other remains fixed with a hybrid reward function, that guides this training by evaluating the final answer's correctness, adherence to conversational format, and overall response length. This iterative self-evolution process enhances sophisticated reasoning capabilities while significantly reducing token usage.

\section{Methodology}\label{sec: methodology}
\subsection{Multi-Round Inference}
We use the multi-round inference model shown in Fig.~\ref{fig: multiround}. In the first round, the LLM is asked to respond to the question in the prompt. Next round onwards, the LLM is fed the original prompt along with the summary answer from the previous round. The system prompt mentions the whole multi round inference process to allow the LLM to think modularly. The original system prompt is the following:

\begin{tcolorbox}[colback=gray!10!white,
                  colframe=gray!80!black,
                  arc=2mm, 
                  boxrule=0.5pt, 
                  left=1mm, right=1mm, top=1mm, bottom=1mm]
\texttt{``You are a helpful assistant. When the user asks a question, you solve it in 3 rounds.
In each round, you first think about the reasoning process of answering and then provide the user with a detailed progress about it.
The reasoning process and the progress are enclosed within <reasoning> </reasoning> and <answer> </answer> tags, respectively.
Therefore, you follow the strict format:\\
<reasoning>
reasoning process here
</reasoning>
<answer>
detailed progress here
</answer>
\\ \\
The User provides this detailed progress as additional context in the next round.
You then respond again with further thinking and further progress.
When the User says that the current round is the final (third) round, you provide an answer inside the answer tags.
You also enclose a final answer in third round in the box: \textbackslash{boxed}\{\}. Only this boxed final answer is used for evaluation.''}
\end{tcolorbox}

From second round onwards, the summary of the previous round $i$ is added to the prompt with the instruction: \texttt{``Progress in round i: ''}. In the final round, the LLM is further instructed as: \texttt{``Current round is the final (third) round. Provide a final answer.''}.

\begin{figure}
    \centering
    \includegraphics[width=\linewidth]{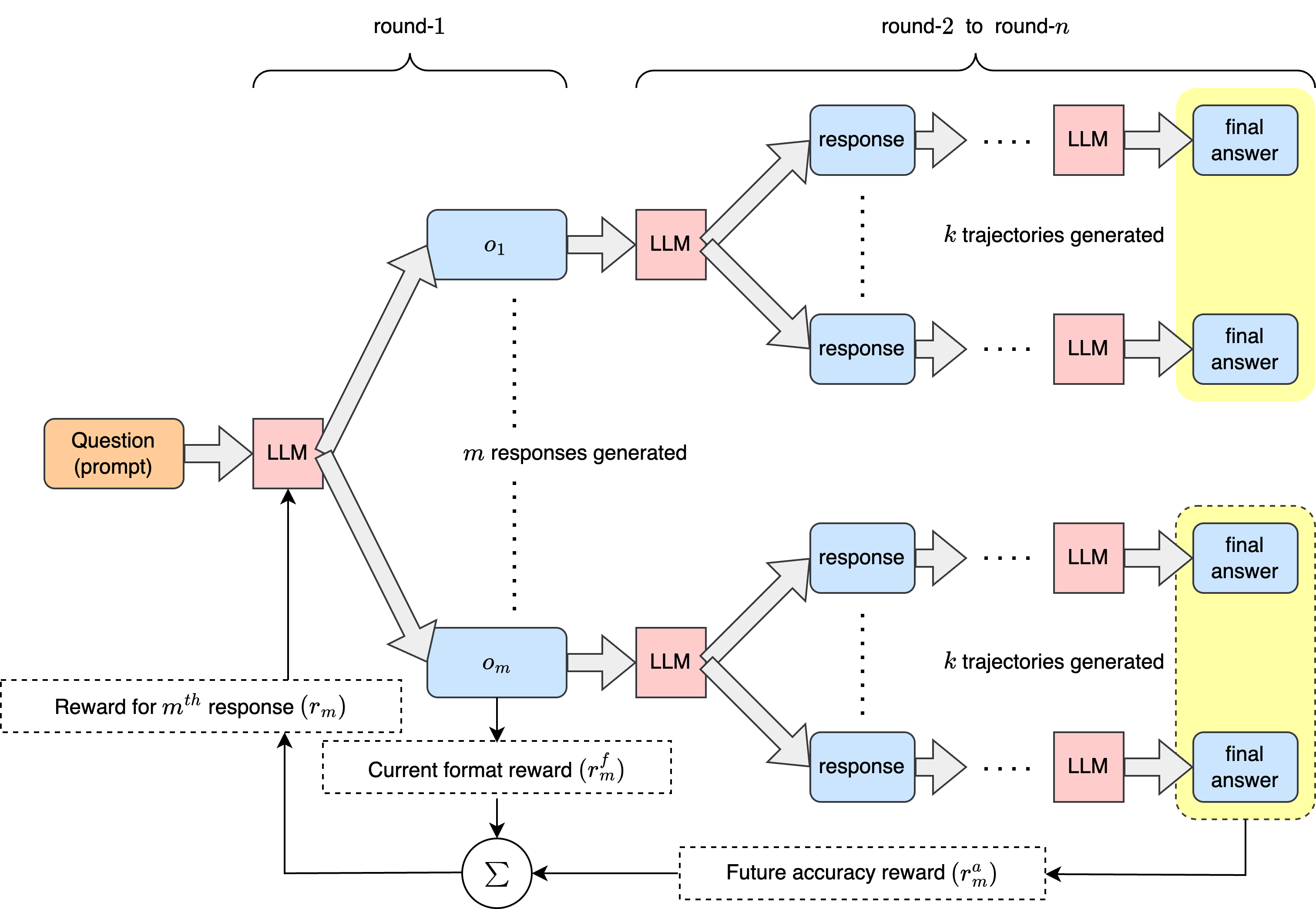}
    \caption{GRPO pipeline of MOTIF algorithm.}
    \label{fig: multiround_grpo}
\end{figure}

\begin{algorithm}[t]
    \caption{MOTIF algorithm using GRPO pipeline}\label{algo: grpo_pipeline}
    \begin{algorithmic}[1]
    \State Initialize model $\pi_{\theta}$ with system prompt, and load question-answer dataset $\mathcal{D}=\{q_j, a_j\}_{j=1}^N$.
    \Procedure{Multi-round-inference}{$\pi_{\theta}$, prompt $= p$, rounds $= R$}
    \State $o = \pi_{\theta}(p)$
    \State $ans = $ extracted\_answer$(o)$
    \For{$r = 2, 3, \cdots, R$}
    \State $o = \pi_{\theta}(\text{augment}(p, ans))$
    \State $ans = $ extracted\_answer$(o)$
    \EndFor
    \State return $ans$
    \EndProcedure
    \Procedure{MOTIF-Training}{$\pi_{\theta}, \mathcal{D}$}
    \For{epoch $= 1, 2, \cdots, E$}
    \For{$j = 1, 2, \cdots, N$}
    \State Temperature is set to $0.8$ for generating diverse set of answers.
    \For{$i = 1, 2, \cdots, m$}
    \For{$\ell = 1, 2, \cdots, k$}
    \State $f^i_{\ell} = $ Multi-round-inference$(\pi_{\theta}, q_j, R)$
    \EndFor
    \State Calculate rewards: $r_i = r^a_i + r^f_i$
    \EndFor
    \State $\pi_{\theta} \leftarrow $ GRPO\_step$(\pi_{\theta}, \{r_i\}_{i=1}^m)$
    \EndFor
    \EndFor
    \State return $\pi_{\theta}$
    \EndProcedure
    \end{algorithmic}
\end{algorithm}

\subsection{Outcome Based Reward Function}
We use GPRO algorithm, proposed by \cite{shao2024deepseekmath} for RL, however, with a crucial difference. The reward function does not calculate the accuracy from the first round of inference. Rather, in the first round $m$ samples are generated. Each of these individual sample answers are then used for further multi-round inference up to the $n$th round. For each first round response, we generate $k$ trajectories. For question and correct answer pair $(q,a)$, we denote the first round outcomes as $\{o_1, o_2, \cdots, o_m\}$. For outcome $o_i$, the corresponding final answers extracted after the $n$th round are denoted as $\{f^i_1, f^i_2, \cdots, f^i_k\}$. Therefore, we formulate the accuracy reward as the estimate of the average accuracy of getting the correct answer after the $n$th round
\begin{align}
    r^a_i = \frac{1}{k}\sum_{j=1}^{k}\mathbb{I}(\text{\textbackslash{boxed}\{}a\text{\} is in } f^i_j),
\end{align}
where, $\mathbb{I}(\cdot)$ is the indicator function. This reward gives an estimate of the probability of getting the correct answer from the outcome $o_i$ as shown in Fig.~\ref{fig: multiround_grpo}. The format reward of the first round, on the other hand, checks whether the correct think and answer tags were followed in the response. We denote the format reward for $i$th response as
\begin{align}
    r^f_i = \mathbb{I}(o_i\text{ is correctly formatted with answer and reasoning tags}).
\end{align}
The total reward, therefore, is $r_i = r^a_i + r^f_i$. Since our objective for RL is to maximize the expected value of this overall reward, we are essentially optimizing the reasoning performance over multiple rounds. Note that, this formulation does not require optimizing for the responses in the intermediate steps. Rather, only the first round outcome is sufficient to assign a reward to. With this rule based reward formulation, the GRPO objective is defined as 
\begin{align}\notag
&\mathcal{J}_{GRPO}(\pi_\theta) \\ \notag
&= \mathbb{E}_{q \sim P(Q), \{o_i\}_{i=1}^m \sim \pi_{\theta_{old}}(\cdot|q)} \\ 
&\Bigg[\frac{1}{m}\sum_{i=1}^m\frac{1}{|o_i|} \sum_{t=1}^{|o_i|} \min \left\{ \frac{\pi_\theta(o_{i,t} | q, o_{i, <t})}{\pi_{\theta_{old}}(o_{i,t} | q, o_{i, <t})} \hat{A}_{i,t}, \text{clip} \left( \frac{\pi_\theta(o_{i,t} | q, o_{i, <t})}{\pi_{\theta_{old}}(o_{i,t} | q, o_{i, <t})}, 1 - \epsilon, 1 + \epsilon \right) \hat{A}_{i,t} \right\}\Bigg],
\end{align}
where $\pi_{\theta}$ and $\pi_{\theta_{old}}$ are current policy and policy before update, respectively. $P(Q)$ is the distribution of the questions in the training data; $\epsilon$ is the clipping hyperparameter; $o_{i,t}$ and $o_{i, <t}$ represent the $t$th outcome token and the tokens before that, respectively.  $\hat{A}_{i,t}$ is the estimated advantage of the $t$th token and is calculated as 
\begin{align}
    \hat{A}_{i,t} = \frac{r_i - \text{mean}(r_1, r_2, \cdots, r_m)}{\text{std}(r_1, r_2, \cdots, r_m)}.
\end{align}
Note that, we have omitted the factor corresponding to $\beta$ in the GRPO formulation, in comparison to the original formulation by \cite{shao2024deepseekmath}. Since we are using rule based reward, the model cannot hack the reward to get the correct answer, and thus we do not need any reference policy to compare the output to. The whole GRPO pipeline is described is Algorithm~\ref{algo: grpo_pipeline}.

\section{Experiments}\label{sec: experiments}
\subsection{Settings}
We implement MOTIF using the GRPO RL setting of \cite{brown2025grpodemo}. We use parameter efficient fine-tuning by \cite{unsloth} to reinforcement fine-tune Qwen2.5-3B-Instruct base model. We choose LoRA rank as 64 to update around 4\% of total model parameters for the fine-tuning. We use GSM8K dataset, which contains grade school level math questions and answers, for generating sample responses. In addition to MOTIF training, we have a vanilla GRPO trained model as a baseline for comparison. The GRPO training uses the following system prompt:

\begin{tcolorbox}[colback=gray!10!white,
                  colframe=gray!80!black,
                  arc=2mm, 
                  boxrule=0.5pt, 
                  left=1mm, right=1mm, top=1mm, bottom=1mm]
\texttt{``You are a helpful assistant.
When the user asks a question, you first think about the reasoning process in mind and then provide the user with an answer.
The reasoning process and the answer are enclosed within <reasoning> </reasoning> and <answer> </answer> tags, respectively.
In your answer, you also enclose your final answer in the box: \textbackslash{boxed}\{\}.
Therefore, you respond in the following strict format:\\
<reasoning>
reasoning process here
</reasoning>
<answer>
answer here
</answer>.''}
\end{tcolorbox}
We also calculate the accuracy of the base LLM to measure the overall improvement using the following system prompt:

\begin{tcolorbox}[colback=gray!10!white,
                  colframe=gray!80!black,
                  arc=2mm, 
                  boxrule=0.5pt, 
                  left=1mm, right=1mm, top=1mm, bottom=1mm]
\texttt{``You are a helpful assistant. In your response, you enclose your final answer in the box: \textbackslash{boxed}\{\}.''}
\end{tcolorbox}

For both GRPO and MOTIF, each question is used for genering 8 sample responses. In MOTIF, for such response, we generate 4 samples of multi-round inference for reward calculation. The total number of rounds is 3. Since this additional inference process uses resource, we use the same wall-clock training time in both, for fair comparison. This leads to using fewer data points in MOTIF. In GRPO, we use 2000 samples of the GSM8K dataset, while for MOTIF, only 300 samples are used, which is 15\% of GRPO. Both trainings were done using a single NVIDIA A100 40GB GPU.

\subsection{Results}
\begin{table}[ht]
  \caption{Pass@1 accuracy comparisons in benchmarks}
  \label{table: results}
  \centering
  \begin{tabular}{lcc}
    \toprule
    \multirow{2}{*}{Model} & \multicolumn{2}{c}{Dataset} \\
    \cmidrule(lr){2-3}
           & MATH500 & AIME2024 \\
    \midrule
    Qwen2.5-3B-Instruct (base model) & 37.6\% & 0.0\% \\
    GRPO training & 44.8\% & 3.33\% \\
    \rowcolor{gray!20}
    MOTIF training & 48.6\% & 6.67\% \\
    \bottomrule
  \end{tabular}
\end{table}

There are multiple ways to measure accuracy of an LLM in benchmarks. For our purposes, we use the simplest pass@1 accuracy~\citep{chen2021evaluating}, which measures the probability of getting a correct answer in the first attempt. It is defined as 
\begin{align}
    \text{Pass@1 accuracy} = \frac{\text{\# correct answers}}{\text{\# questions}} = \frac{1}{L}\sum_{j=1}^{L}\mathbb{I}_j, 
\end{align}
where $L$ is the total number of questions in the benchmark; $\mathbb{I}_j = 1$, if the boxed answer to the $j$th question is in the response from the model, and $0$, otherwise.

The results of our experiment are shown in Table~\ref{table: results}. As we observe, MOTIF shows 3.8\% and 3.3\% improvements over vanilla GRPO in MATH500 and AIME2024 benchmarks, respectively. Both the training significantly improves the performance of the base model. However, MOTIF only uses 15\% of the training samples to do so.

\begin{figure}[t]
  \centering
  \begin{subcaptionbox}{Reward over training steps.\label{fig: motif_reward}}[0.46\linewidth]
    {\includegraphics[width=\linewidth]{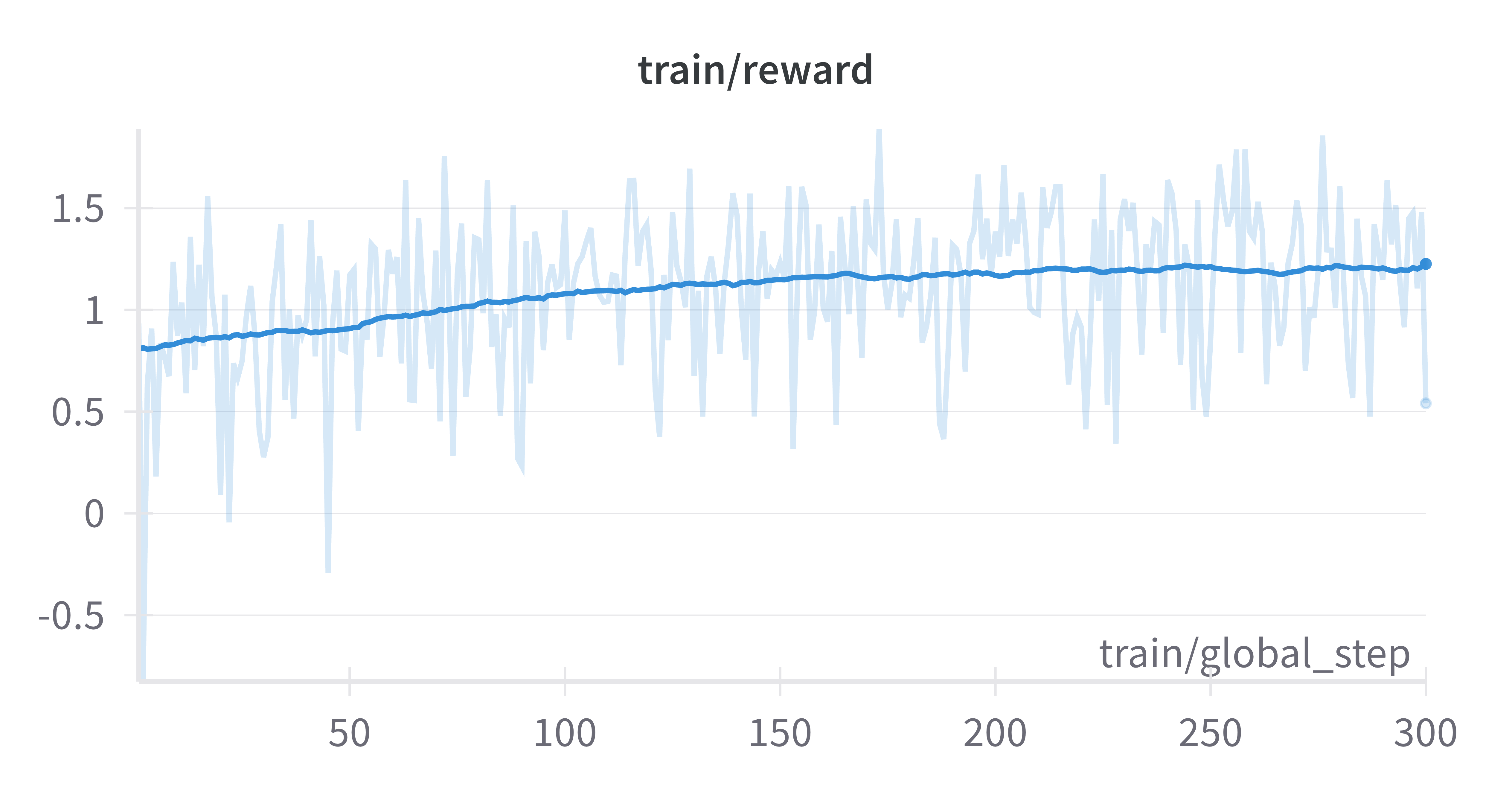}}
  \end{subcaptionbox}
  \hspace{0.05\linewidth}
  \begin{subcaptionbox}{Completion  length over training steps.\label{fig: motif_length}}[0.46\linewidth]
    {\includegraphics[width=\linewidth]{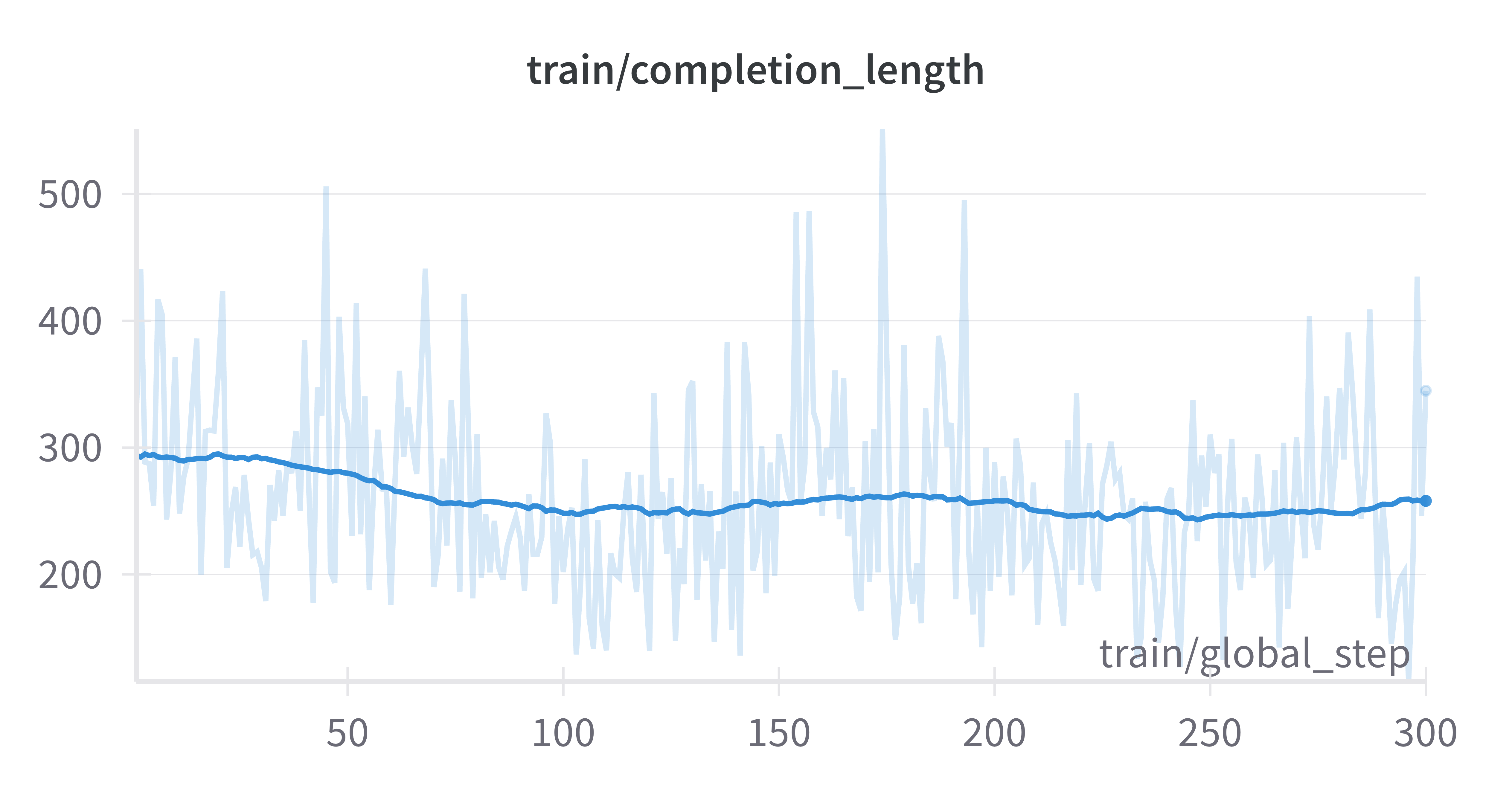}}
  \end{subcaptionbox}
  \caption{Training evolution in MOTIF shows that the model trains fast, even with very small number of training samples, while the overall completion length does not increase much.}
  \label{fig: motif_train}
\end{figure}

We observe further details of training in Fig.~\ref{fig: motif_train}. In Fig.~\ref{fig: motif_reward}, we notice that the expected reward grows from the beginning and smoothly over the training steps. This contrasts with regular GRPO training, where the reward takes a large number of steps to grow significantly. Another observation from Fig.~\ref{fig: motif_length} is that the average response length actually decreases a little over the training steps. This also contrasts with the regular GRPO training, where the response length increases as the training progresses. This indicates that, as the overall response gets distributed over multiple rounds, each of the individual rounds only performs a fraction or a module of the reasoning for the total answer.

\section{Conclusion}\label{sec: conclusion}
In this work, we propose MOTIF -- a multistep reinforcement fine-tuning method that enables an LLM to think modularly, i.e., thinking and answering over multiple rounds. We show that outcome based reward can be formulated for training such a system with verifiable rewards. We employed a GRPO pipeline to train open source LLM: Qwen2.5-3B-Instruct on GSM8K dataset. We tested the pass@1 accuracy of the trained model benchmarks, like MATH500 and AIME2024. Our experiments showed significant improvements in accuracy and training time over vanilla GRPO training in both the benchmarks. In MATH500, MOTIF performed 3.8\% better  and in AIME2024, 3.3\% better than vanilla GRPO algorithm. Furthermore, this improvement was achieved by utilizing only 15\% of samples used in GRPO. This indicates that MOTIF is a much more sample efficient training algorithm, compared to vanilla GRPO training.


\bibliography{references}  

\end{document}